\documentclass[journal]{IEEEtran}
\usepackage[utf8]{inputenc}
\usepackage{graphicx}

\usepackage{tabularx}

\newcommand\clearrow{\global\let\rowmac\relax}

\usepackage{graphicx}
\usepackage[inline]{enumitem}

 \usepackage{cite}
\usepackage{subcaption}
\usepackage{hyperref}

\usepackage[font=small,labelfont=bf]{caption}

\usepackage{multirow}

\usepackage{fourier} 
\usepackage{array}
\usepackage{makecell}
\usepackage{subcaption}

\begin{document}

\title{Short-Term Load Forecasting using Bi-directional Sequential Models and Feature Engineering for Small Datasets}

\author{Abdul Wahab, Muhammad Anas Tahir, Naveed Iqbal, Faisal Shafait~\IEEEmembership{Member,~IEEE}, and Syed Muhammad Raza Kazmi~\IEEEmembership{Member,~IEEE}}




\maketitle

\begin{abstract}
 Electricity load forecasting enables the grid operators to optimally implement the smart grid's most essential features such as demand response and energy efficiency. Electricity demand profiles can vary drastically from one region to another on diurnal, seasonal and yearly scale. Hence to devise a load forecasting technique that can yield the best estimates on diverse datasets, specially when the training data is limited, is a big challenge. This paper presents a deep learning architecture for short-term load forecasting based on bidirectional sequential models in conjunction with feature engineering that extracts the hand-crafted derived features in order to aid the model for better learning and predictions. In the proposed architecture, named as Deep Derived Feature Fusion (DeepDeFF), the raw input and hand-crafted features are trained at separate levels and then their respective outputs are combined to make the final prediction. The efficacy of the proposed methodology is evaluated on datasets from five countries with completely different patterns. The results demonstrate that the proposed technique is superior to the existing state of the art.
 
\end{abstract}

\begin{IEEEkeywords}
Short Term Load Forecasting, Small Datasets, Deep Learning, Feature Engineering, Bi-directional Sequential Models, Deep Derived Feature Fusion (DeepDeFF).
\end{IEEEkeywords}

%
\IEEEpeerreviewmaketitle

\section{Introduction} 

Smart grid, in simple terms, implies monitoring and control of the power system's assets in the generation, transmission, distribution, and utilization, to achieve high efficiency and reliability at low operational costs. Its cardinal feature -- demand response, can only be fully realized through accurate forecasting of various variables, the most important of which is the forecasting of electrical load~\cite{lfdr}. Artificial intelligence is fast becoming cardinal for data analytics and enhanced control of modern power systems. One of its most desirable application in recent times is the load forecasting through machine learning for predicting the trends in energy demand, so that the control decisions can be proactively optimized. 

Long-term~\cite{ltlf}, mid-term~\cite{mtlf}, and short-term~\cite{stlf} are the different types of load forecasting found in the literature based on their duration of prediction from years to minutes. Short-term load forecasting (SLF) is more difficult than the mid and long term forecasting because of the greater variance in the respective energy consumption patterns~\cite{smart-grid-aus}. The advantage of SLF is that it provides better insight into the electricity consumption patterns and a greater degree of freedom for demand-side management. Also, SLF can be aggregated to get mid-term and long-term forecasts. Therefore this paper focuses on SLF.

One of the most challenging problems in SLF is posed by small datasets; particularly in the case of individual households which usually exhibit wide variations in energy consumption in short intervals, thus making it harder for the deep learning models to learn the underlying patterns~\cite{smart-grid-aus}. Before the deep learning era, lot of research went into the engineering of hand-crafted features which were required as the inputs to machine learning algorithms. The capability of deep learning to extract implicit features, removed the need of such complicated pre-processing of raw data. However deep learning models require large training data to extract useful features, more so for the datasets with high variances.

This paper presents a novel deep learning architecture that combines the use of hand-crafted features with raw data, such that the deep learning model can work well for SLF of small datasets. The results demonstrate significant improvements in the performance, specially for small datasets by using the proposed architecture.


\section{Literature Review}
A significant amount of research has been carried out in recent times to develop SLF as the enabling tool for efficient monitoring and control of power system. Before the deep learning era, hand-crafted features were used to be fed to a machine learning model for making predictions. In~\cite{santos-martins} feature engineering was done to design a feature vector by performing entropy analysis with a specific tolerance band and auto-correlation function. The designed feature vector was then passed through an artificial neural network (ANN) for prediction. Ferreira and da Silva used a Bayesian based approach to solve the complexity of neural network and variable selection~\cite{ferreira-silva}. The approach has theoretical ground but relies on various assumptions regarding the network parameters distribution, requires three relevance thresholds and is computationally expensive. Phase-space embedding method was used for the selection of input variable which allowed to include the preference of the past values of prediction quantity in the input vector~\cite{drezga-rahman}. A neural networks based approach to forecast next 24 hour load on medium and low voltage substations was presented in~\cite{ding-benoit}. The use of separate models each for daily average power and for intraday variation in power, improved the accuracy of prediction compared to the model based on time series.


Learning the daily routine of the usage of various appliances can help in better forecasting of an household's load profile. It was shown in ~\cite{ampd-kong-ca, 8282478} that using the consumption data of the appliances together with the aggregated data of the whole house as the input to the long short term memory (LSTM) models gave better results than using the whole house readings alone.

Recently recurrent neural networks (RNN) have become the popular choice for load forecasting. In~\cite{theile2018day} machine learning models were used for predicting the energy demand on publicly available RTE dataset~\cite{rte-data}. The performances of RNN and support vector machine (SVM) models were compared using different input features. The models were evaluated on a test set of 10 days of year 2017. The results demonstrated that RNN performed better, with a MAPE of 3.52\%, compared to SVM with a MAPE of 14.00\%. 

A recent study~\cite{smart-grid-aus} demonstrated how the individual household level load forecasting can be challenging because of different patterns of energy consumption of individual consumers~\cite{smart_grid}. A two layers based LSTM model was proposed and compared with other models based on back-propagation neural network (BPNN), \textit{k}-nearest neighbour (KNN), extreme learning machine (ELM) and input scheme combined with a hybrid forecasting framework (IS-HF). Individual models for each household were trained and the best average MAPE of 44.06\% was achieved through LSTM. Also it was demonstrated that aggregating these individual forecasts resulted into quite the same net MAPE of $\sim$8\% that was yielded when the aggregated data of the consumers was trained and tested on a single model. This difference in the MAPE of individual versus aggregated forecast established that the individual SLF is harder compared to aggregated one; however the advantage of individual SLF is that it provides better insight into the trend of each constituent customer and can easily be aggregated together to provide the net trend.

Electricity demand is influenced by weather, holiday, time of day etc. Time dependant convolution neural network (TD-CNN) and cycle based long short term memory (C-LSTM) for short- and medium-term load forecasting was presented in~\cite{han-lingyi}. Electric load on weekly basis was arranged in image format on which TD-CNN was run. C-LSTM helped to extract time dependencies between sequences. The models performed better than the traditional LSTM model while reducing the training time.

Another important application of SLF is in energy trading which is a complex process due to non-periodic variations in energy consumption. Accurate forecasting for hourly spot price is the key to achieve the best trading decision which is vital for investors and retailers in electricity market. A model based on a hybrid approach comprising of ARIMA, multiple linear regression (MLR), and Holt-Winter model was proposed in~\cite{bissing-daniel}. The hybrid model was tested for Iberian electricity market dataset to forecast hourly spot prices for various numbers of days. A hybrid model based on non-linear regression and SVM was proposed in~\cite{ziming2018month}, that was tested on ERCOT data~\cite{ercot-data}. This hybrid model achieved MAPE of 7.30\% compared to the individual models with 8.99\% and 8.63\% MAPE respectively.

Improvement of forecasting accuracy using standard LSTM model by feeding it processed features rather than raw data was proposed in~\cite{8996379}. The power load sequence was decomposed by complementary ensemble empirical mode decomposition (CEEMD), then the approximate entropy (AE) values of the obtained subsequences were calculated. The subsequences with similar AE values were merged into new sequence to form the inputs of the load forecasting model. This reduced the complexity of the power load sequence and improved the accuracy of load forecasting. The vanilla LSTM network was improved in~\cite{8833755} by cleaning and processing the raw load data using isolated forest algorithm.

Electric load forecasting requires training of large number of neurons in hidden layer, which increases the size of the network and slows overall training process. To reduce this overhead, a multi-column radial bias function (MCRF) with error correction algorithm designed to reduce the number of hidden neurons in a network, was proposed in~\cite{8835054}. It was shown that MCRF with only 50 neurons in hidden layer took only 10 minutes to train and achieved the MAPE of 4.59\% compared to other models with more than 150 neurons that achieved better MAPE of 1.77\% but took hours to train. 

Accuracy of SLF can be improved through careful analysis of the load data to find the effectiveness of selected features. A technique was proposed in~\cite{8890630} for features selection where the bisecting K-means algorithm was used to cluster the load data with high similarity for a forecast date. The ensemble empirical mode decomposition (EEMD) helped to combine components with similar entropy. A bidirectional recurrent neural network (BRNN) model was proposed to forecast the load of the network. The model was verified on two datasets including a dataset from load forecasting competition. The results showed that BRNN model performed better even than the winner of the competition.








The literature survey therefore implies that a better load forecasting technique with reduced statistical error is a hot topic of research for modern power systems.

\section{Proposed Methodology}
Recently~\cite{han-lingyi, smart-grid-aus, ampd-kong-ca} deep learning solutions, particularly sequential models such as RNN and LSTM models are becoming popular choices for load forecasting. LSTM~\cite{Sepp-Hochreiter} has become a state of the art tool for time series problems owing to its ability to learn temporal patterns in sequential data. This paper presents a novel architecture named as deep derived feature fusion (DeepDeFF), comprising of a bidirectional sequential model with feature engineering for realizing a more accurate SLF technique.

\subsection{Bidirectional Sequential Models}

Bidirectional model trains forward and reverse nodes using respectively: 
\begin{enumerate*}
    \item input in positive time, i.e. the given input as it is,
    \item input in negative time, i.e. a time-reversed copy of the original input. 
\end{enumerate*}
The advantage of bidirectional model compared to conventional ANN models is that it observes the input in both forward and reverse directions to extract more information from the input sequence. This technique of negative time and bidirectional layer was first discussed in~\cite{mike_schuster_blstm}. 


This paper implements LSTM, RNN, gated recurrent unit (GRU) as well as their bidirectional counterparts (BLSTM)~\cite{blstm}, BRNN and (BGRU) on several datasets for a comprehensive comparison presented in the results.

\subsection{Derived Features}


The aim of derived features is to enrich the training data with useful features for more accurate predictions. A deep learning model with enough computation time and data may extract such features on its own, but this cannot be guaranteed within the constraints of time and resource. Thus providing these derived features explicitly as inputs can enable the model to learn more from the data and converge quickly, specially for small datasets. Generally the performance of deep learning models improve by increasing the number of relevant input features unless it starts to over fit.

The basic features used for generation of derived features and as input for the DeepDeFF model are:

\begin{itemize}
\item Energy load consumption $E$.

\item Time-stamp of the day $T$, divided into 30 minutes interval each. The feature is converted into One-hot encoding.

\item Current day of the week $W$, converted into one-hot encoding.

\item Holidays represented by a binary label $H$. At the moment only weekends are marked as holidays, but in future work this can be expanded and synced with other public and national holidays.



\end{itemize}

Derived features are calculated for each record in the input sequence (\textit{1}, $K$, $f$), where $K$ represents the number of past records used for creating the input sequence and $f$ represents the basic features. Following are the derived features that are calculated and used as input to the DeepDeFF model. 
\begin{itemize}
\item Average load consumption of $K$ time-steps.

\item Standard deviation of load consumption of $K$ time-steps.

\item Average load consumption of the time-stamp $t$ that is to be predicted, for past $K$ days.

\item Standard deviation of load consumption of the time-stamp $t$ that is to be predicted, for past $K$ days.



\end{itemize}

\subsection{Proposed Architecture}

This paper proposes a two-layer bidirectional sequential model architecture DeepDeFF, which inputs the raw and derived input features into separate layers to extract learned features. The idea behind using separate input layers for basic and derived sequences is to allow the sequential layers to learn from the two input sequences independently. The goal is to exploit the relevance of basic and derived sequences with the predictions individually. The learned representation from the individual sequential layers is then merged and fed to a dense layer followed by a linear activation output layer to make the final prediction of the load at the next time interval.

\begin{figure*}[htp]
    \centering
    \includegraphics[width=1.0\textwidth]{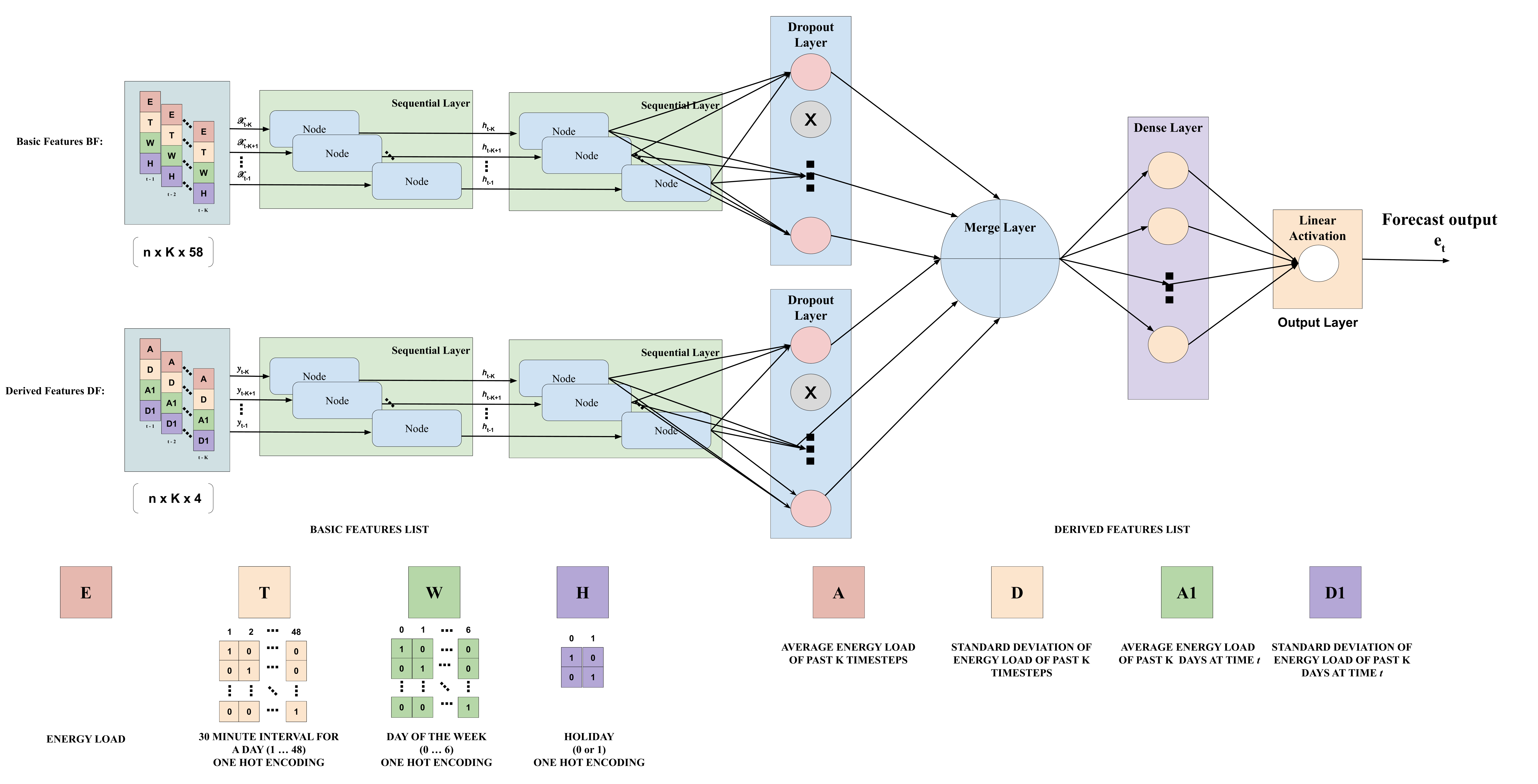}
    \caption{System Architecture: The input data is first pre-processed to achieve the derived input features. The raw input features $T$, $W$, $H$ are converted to one-hot encoding. The raw and derived input features are then fed to two individual bi-directional sequential layers. The information extracted from these separate layers is then merged and used as input to a dense layer followed by a final feed forward layer with $Linear$ activation function.}
    \label{fig:architecture}
\end{figure*}

Fig.~\ref{fig:architecture} shows the schematics of the DeepDeFF architecture. The hyper parameter settings consists of:
 \begin{enumerate}[label=(\alph*)]
     \item 20 nodes sequential layer
     \item a dropout of 0.2
     \item Adam optimizer
     \item MAPE as loss function
     \item Learning rate: 0.01
\end{enumerate}

\section{The Datasets}
\label{datasets}
The proposed methodology for SLF has been evaluated on five energy load datasets from different sources. This section provides the salient parameters of the dataset and presents the pre-processing technique adopted for each. 

\subsection{Smart Grid Smart City (SGSC) dataset}
SGSC project was initiated by the Australian Government in year 2010~\cite{smart_grid}. It gathered smart-meter data from around 78,000 customers for a period of 4-years. In~\cite{smart-grid-aus}, individual models for each customer was proposed. However since it is not feasible to train individual models for $\sim$78,000 customers, therefore 69 customers having “hot water system” were selected. The same subset is extracted here to evaluate the DeepDeFF architecture. 







\subsection{The Almanac of Minutely Power dataset (AMPds)}
AMPds~\cite{ampd} contains electricity, water and natural gas measurements of a single Canadian household with 19 appliances, recorded for 1 year with 1 minute resolution, which is down-sampled to 30 minutes resolution~\cite{ampd-kong-ca}. The variables for raw features used here are the same as for SGSC except that $E$ here is assigned to the Ampere reading.






\subsection{Réseau de Transport d'Électricité (RTE) France dataset}

RTE dataset~\cite{rte-data} is also used here to evaluate the proposed technique. The dataset used spans from year 2013 to 2016 with the sampling interval of 30 minutes. The raw inputs are programmed with same variables as for SGSC above.






\subsection{The Electric Reliability Council of Texas (ERCOT) dataset}
ERCOT dataset~\cite{ercot-data} provides real time and historical statistics surrounding independent system operator (ISO) operations of the Texas region for a period of $\sim$5 years recorded every 1 hour. The raw features variables used here are the same as for SGSC except that the time $T$ here ranges 1-24 since the resolution is 1 hour.







\subsection{Pakistan Residential Electricity Consumption (PRECON)}
PRECON dataset~\cite{Nadeem:2019:PPR:3307772.3328317} records the electricity consumption patterns in a developing country for 42 households of varying financial status, daily routine and load profile. The data is collected with 1 
minute interval from 01-06-2018 to 31-09-2019. The amount of data varies for each household due to different number and types of appliances that are selected for monitoring. This dataset also captures the problem of power outages rampant in developing countries. This is evident from several long 0KW data intervals. For raw features, same variables as in SGSC are used here except that $E$ here refers to the KW usage.

\section{Experiments \& Results}
\label{results}

The proposed framework for SLF is achieved through an evolutionary process after numerous rigorous experiments on all five datasets. This section discusses these experiments in sufficient detail and infers the results obtained. The results from the DeepDeFF architecture are compared with the results of simple two layer sequential models trained on basic features and MAE as loss as proposed in~\cite{smart-grid-aus}.

\subsection{SGSC Dataset}

\subsubsection{Train \& Test Setting}
\label{sgsc_train_test_setting}

The same settings provided in~\cite{smart-grid-aus} are used to extract the subset of SGSC data for fair comparison on the same test set. The data spanning the whole winter season of New South Wales Australia is subdivided into a split ratio of 0.7/0.2/0.1 as:

\begin{enumerate}[label=(\alph*)]
    \item Training set (01-Jun-2013 to 05-Aug-2013)
    \item Validation set (06-Aug-2013 to 22-Aug-2013)
    \item Test set (23-Aug-2013 to 31-Aug-2013)
\end{enumerate}

The first set is to train the DeepDeFF model, validation set is used to select the best model weights based on performance on validation set, while the test set is for the evaluation of the DeepDeFF model. The data is spaced between 30 minutes interval; so for 69 customers the 9 days of evaluation implies the forecasting of 29,808 time points.

\begin{figure*}[htb]
      \centering
\begin{subfigure}{0.48\textwidth}
\includegraphics[width=1\textwidth]{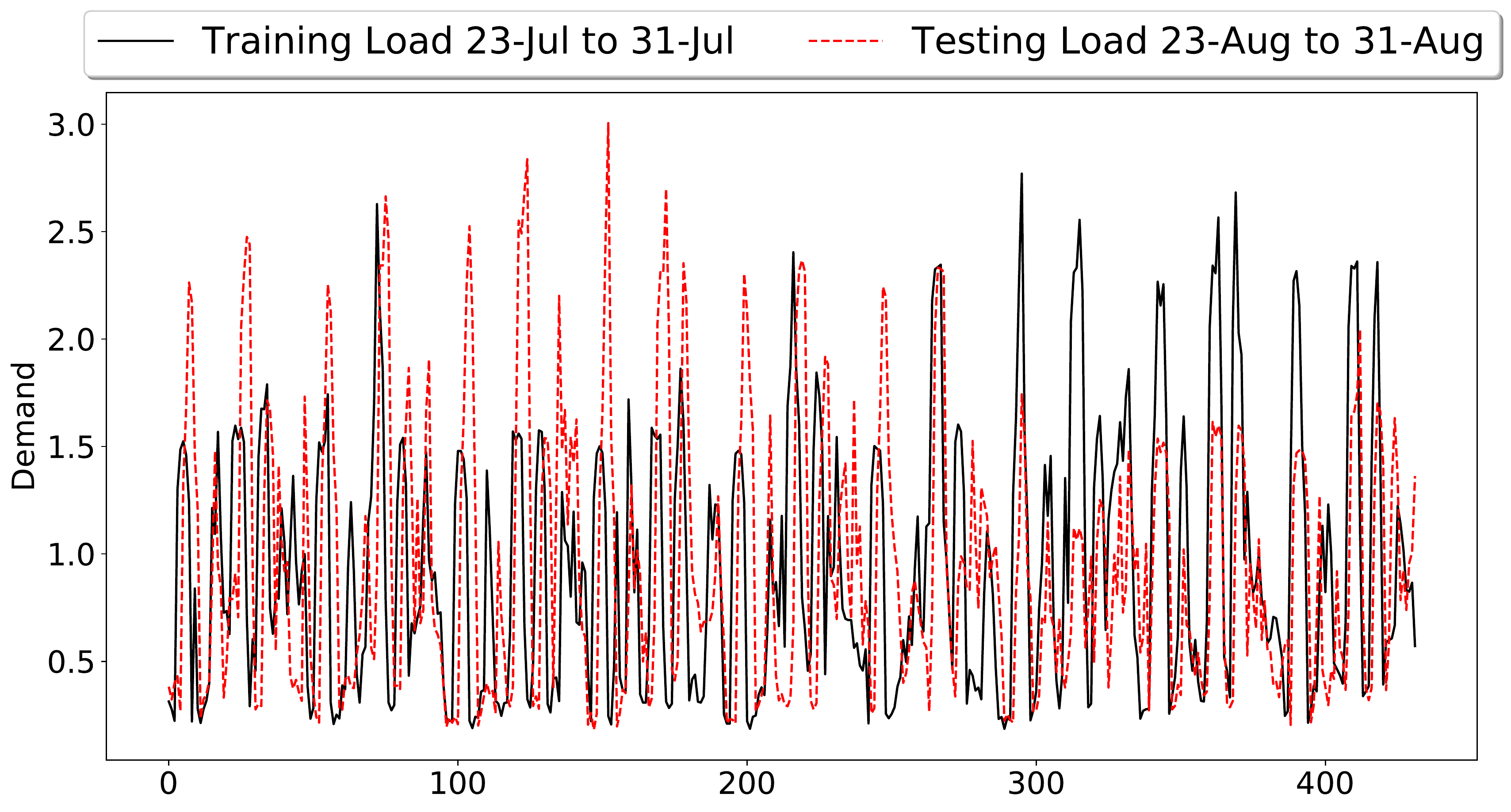}
\caption{Similar train and test pattern for customer 8804804}
\label{good_customer_same_dates}
\end{subfigure}
%
\begin{subfigure}{0.48\textwidth}
\includegraphics[width=1\textwidth]{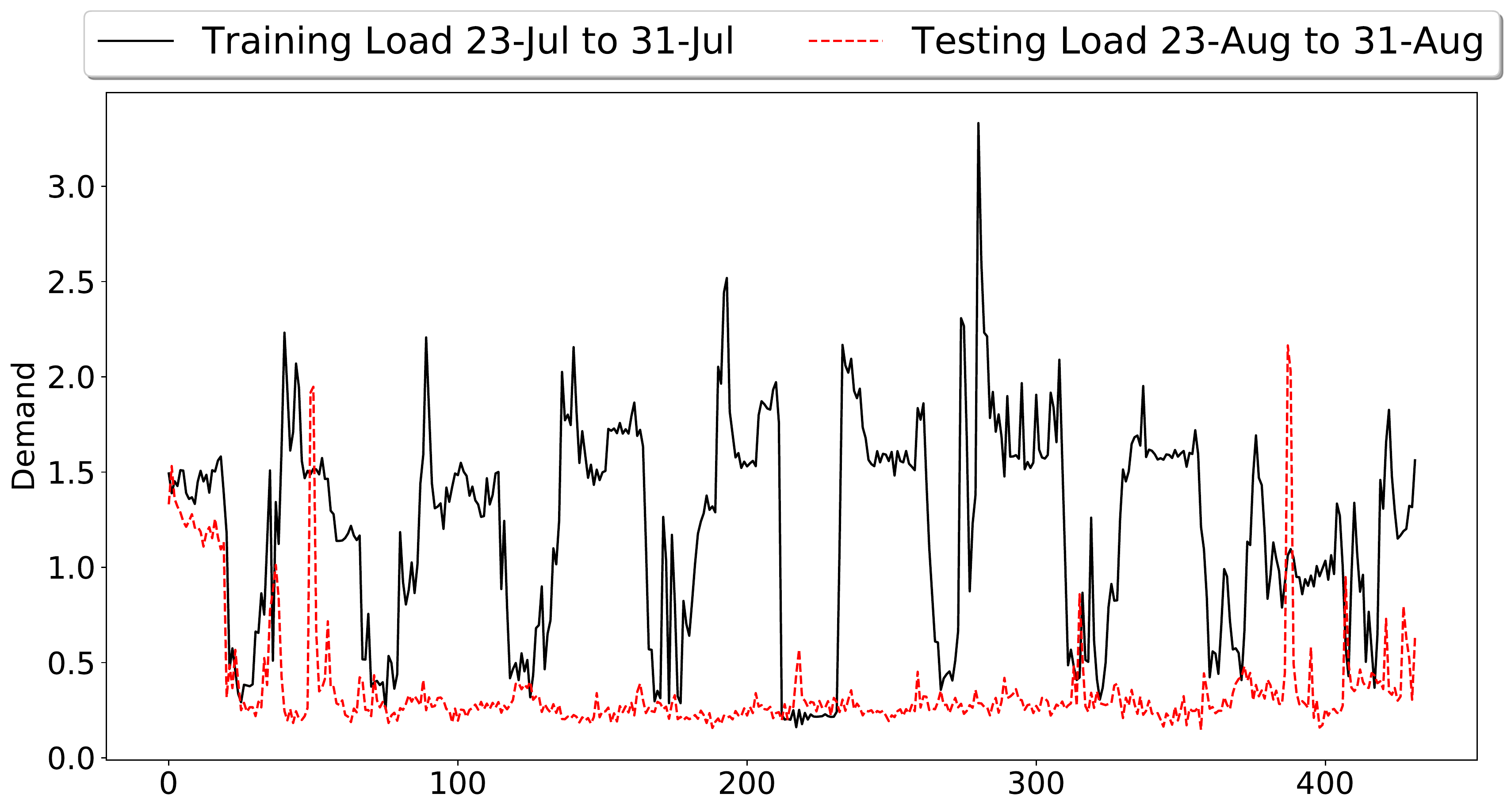}
\caption{Dissimilar train and test pattern for customer 8655993}
\label{bad_customer_same_dates}
\end{subfigure}
%

\caption{A comparison of train and test load data for two types of customers from SGSC dataset is shown, (a) shows a customer whose train and test data patterns have high correlation, (b) shows a customer who has very different load patterns for train and test data.}
\label{sgsc_train_vs_test}
\end{figure*}

Fig.~\ref{sgsc_train_vs_test} provides some insight into the diversity of customers by showing the similarity between their train and test data. Test data is plotted over training data with matching numeric dates. Fig.~\ref{good_customer_same_dates} shows the data of a customer with similarity between train and test data patterns, whereas Fig.~\ref{bad_customer_same_dates} shows no similarity for another customer. This indeed effects the results of DeepDeFF architecture which is reflected in their respective MAPE of 26.04\% and 50.78\% using BLSTM layer; thus the DeepDeFF model has been able to learn the underlying patterns and temporal relations for Fig.~\ref{good_customer_same_dates} but not as good for Fig.~\ref{bad_customer_same_dates}.

\subsubsection{Results}
Table-\ref{sgsc_experiments} shows the comparison of results from rigorous experiments that are performed on SGSC dataset using the proposed DeepDeFF method in contrast with the implementation of the model proposed in~\cite{smart-grid-aus}. The addition of derived features in the proposed architecture along with MAPE as loss function, outperforms the state of the art on the SGSC dataset as evident from the average MAPE computed in Table-\ref{sgsc_experiments}.


\begin{table}[htpb]
\renewcommand{\arraystretch}{1.2}

\caption{Results achieved on SGSC dataset}
\label{sgsc_experiments}
\centering
\begin{tabular}{|c|c|c|c|}
\hline
\thead{Method} & \thead{Time-steps} 
    & \multicolumn{2}{c|}{\thead{Avg.MAPE \%}}
    \\ \cline{3-4} &&
	\multicolumn{1}{c|}{\thead{DeepDeFF}}
	&
	\multicolumn{1}{c|}{\thead{Basic}}\\\cline{3-4}

\hline
\bfseries{BGRU} & \bfseries{2} & \bfseries{34.87}& 42.83\\
\hline
GRU & 2 & 35.01 & 42.85 \\
\hline
BLSTM & 2 & 35.40 & 43.04 \\
\hline
LSTM & 2 & 36.88 & 43.21 \\
\hline
RNN & 2 & 35.94 & 42.04 \\
\hline
BRNN & 2 & 35.79 & 42.74 \\

\hline
BGRU & 6 & 36.02 & 42.08 \\
\hline
GRU & 6 & 35.72 & 41.78 \\
\hline
BLSTM & 6 & 36.41 & 42.96 \\
\hline
LSTM & 6 & 37.28 & 42.41 \\
\hline
RNN & 6 & 38.97 & 42.28 \\
\hline
BRNN & 6 & 38.60 & 42.61 \\

\hline
BGRU & 12 & 36.48 & 42.40 \\
\hline
\bfseries{GRU} & \bfseries{12} & 36.34 & \bfseries{41.54} \\
\hline
BLSTM & 12 & 37.74 & 43.88 \\
\hline
LSTM & 12 & 38.09 & 42.46 \\
\hline
RNN & 12 & 40.57 & 43.29 \\
\hline
BRNN & 12 & 39.64 & 43.20 \\
\hline
\end{tabular}
\end{table}

\begin{figure}[!tbh]
      \centering
      \label{sgsc_results_graphs}
\begin{subfigure}{0.48\textwidth}
\includegraphics[width=1\textwidth]{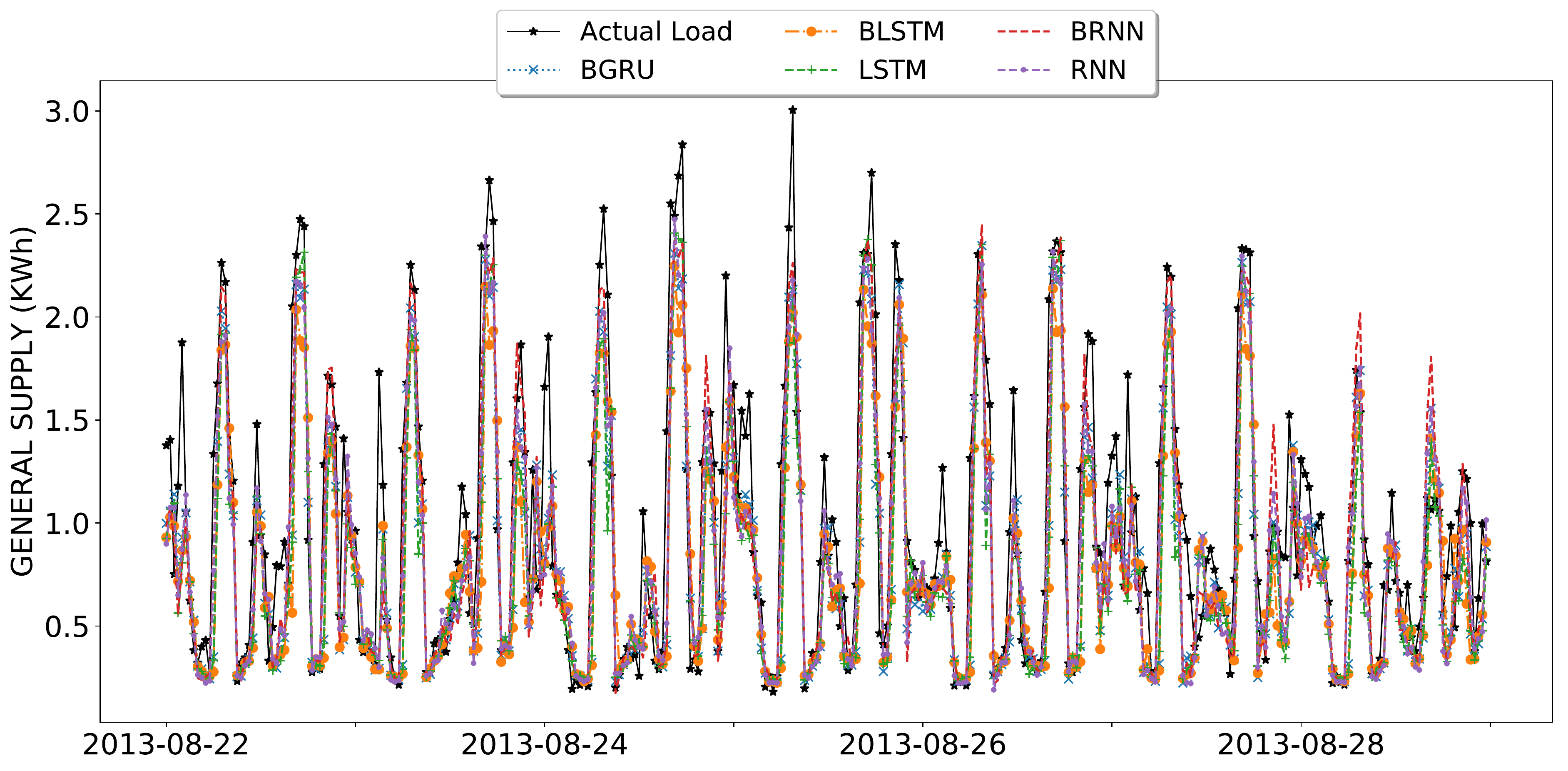}
\caption{Predicted vs. actual data for customer 8804804}
\label{good_customer}
\end{subfigure}
\hfill
\begin{subfigure}{0.48\textwidth}
\includegraphics[width=1\textwidth]{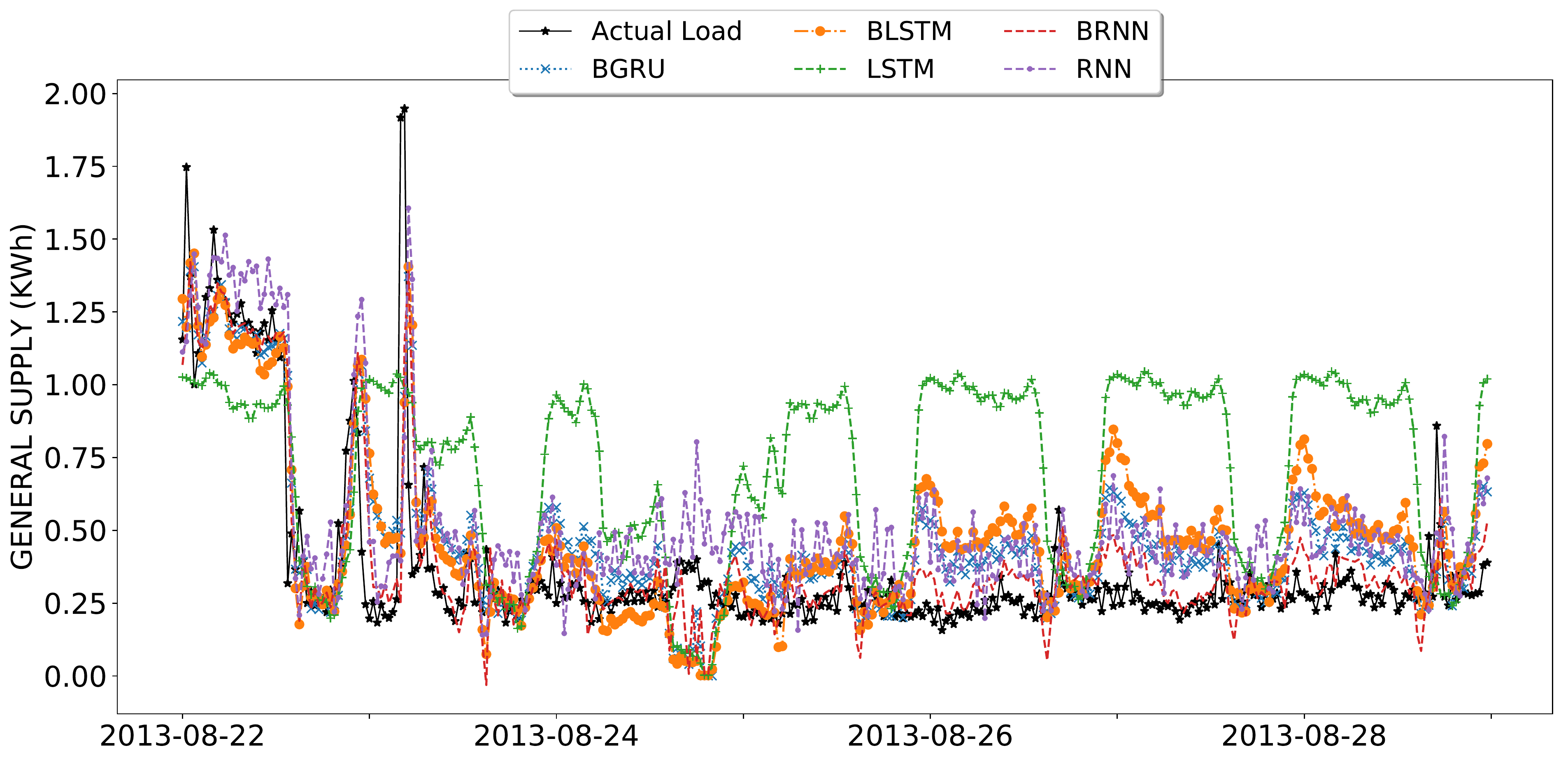}
\caption{Predicted vs. actual data for customer 8655993}
\label{bad_customer}
\end{subfigure}
\caption{Actual load versus load predicted by DeepDeFF architecture utilizing different sequential layers for the two customers shown in~Fig.\ref{sgsc_train_vs_test}}
\end{figure}

Fig.~\ref{good_customer} confirms that the DeepDeFF model indeed predicts the actual load very well for customer 8804804. However, Fig.~\ref{bad_customer} shows that the model under performs for customer 8655993 due to uncorrelated train and test data, owing to disjoint customer behavior during training and testing days.


\subsection{AMPds}
\subsubsection{Train \& Test Setting}
The AMPds data is converted from 1 minute resolution to 30 minutes, yielding 17,483 data points~\cite{ampd-kong-ca}. The data is subdivided with a split ratio of 0.7/0.2/0.1 into:
\begin{enumerate}[label=(\alph*)]
    \item Training set (01-Apr-2012 to 17-Dec-2012)
    \item Validation set (18-Dec-2012 to 23-Feb-2013)
    \item Test set (24-Feb-2013 to 01-Apr-2013)
\end{enumerate}

\begin{figure}[htb]
      \centering
\begin{subfigure}{0.48\textwidth}
\includegraphics[width=1\textwidth]{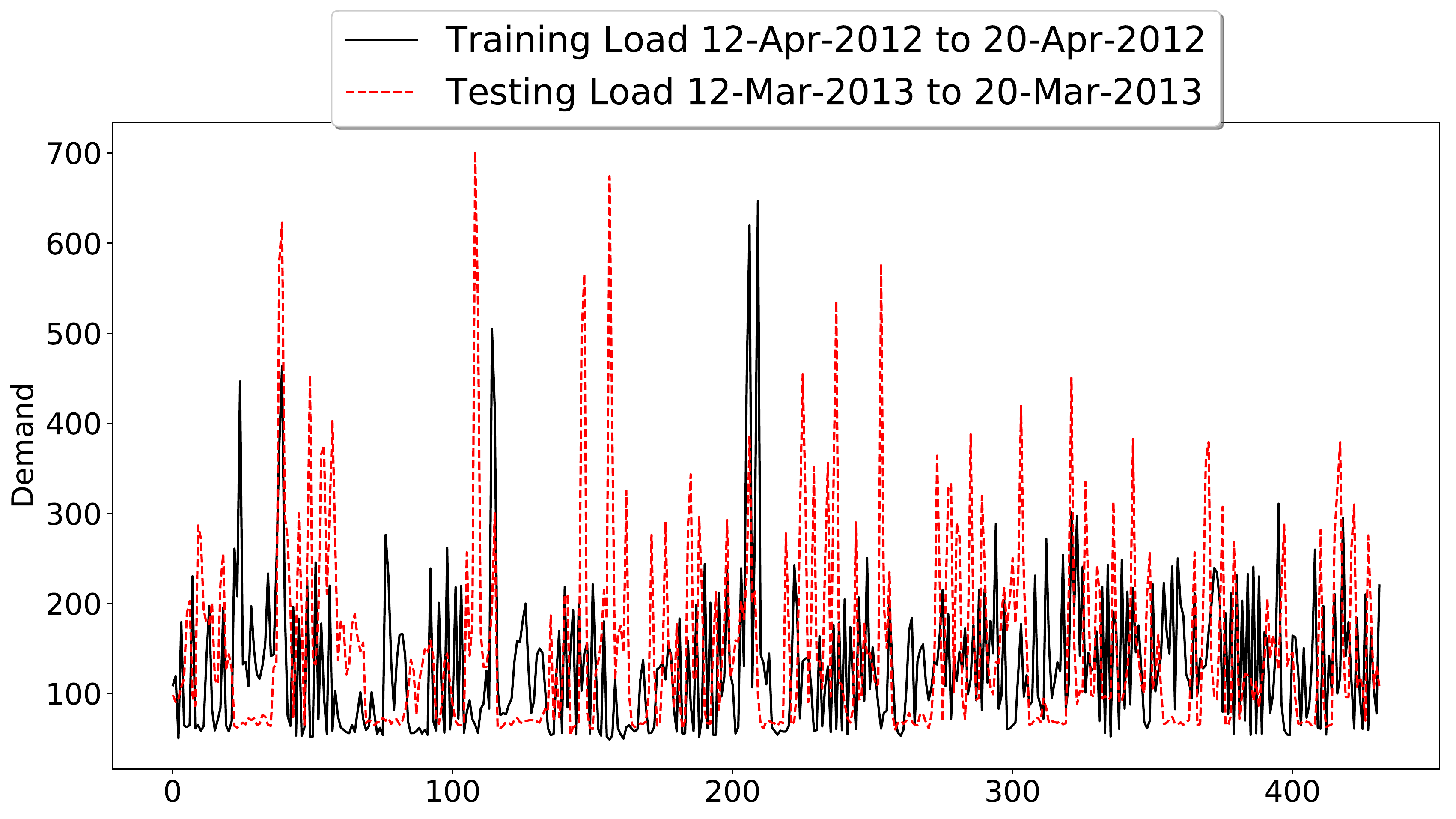}
\caption{Same dates of a month}
\label{ampds_same_date}
\end{subfigure}
\begin{subfigure}{0.48\textwidth}
\includegraphics[width=1\textwidth]{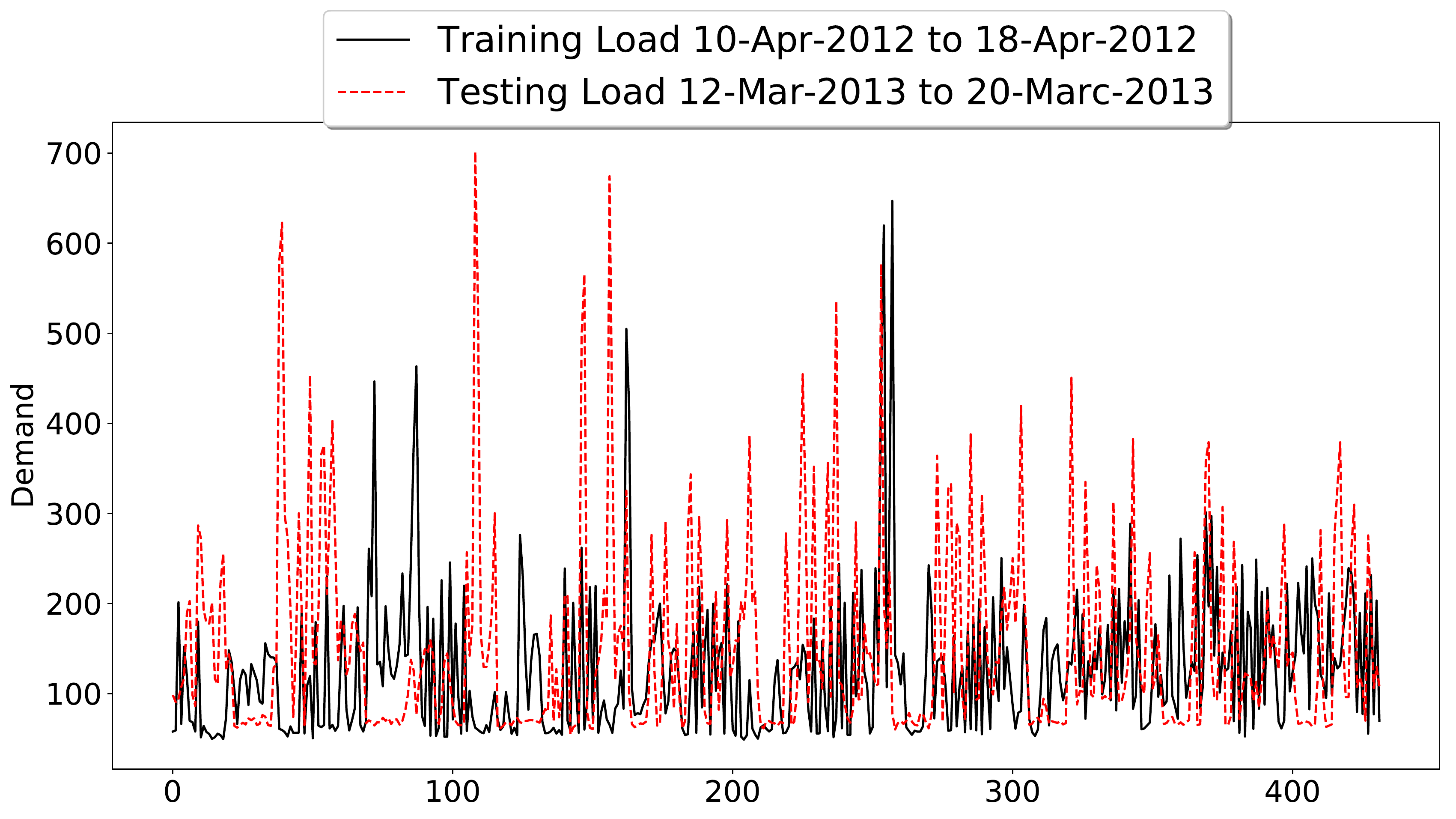}
\caption{Same week days of a month, nearest to the test dates}
\label{ampds_same_days}
\end{subfigure}
\caption{A comparison of train and test data, AMPds}
\label{ampds_train_vs_test}
\end{figure}

Fig.~\ref{ampds_train_vs_test} shows that the train and test data have some common pattern and that there are no abrupt changes like in Fig.\ref{bad_customer_same_dates}. Even though, the training and testing data is not available for same dates of different years, there is a general trend that is being followed in the test data. 

\subsubsection{Results}
Table-\ref{ampds_experiments} shows a comparison of the results produced by the simple two layer sequential model and the DeepDeFF architecture with derived features. The proposed architecture beats the benchmark of 26.23\% achieved in~\cite{ampd-kong-ca} for 6 time-steps.

\begin{table}[htb]
\renewcommand{\arraystretch}{1.2}

\caption{Results achieved on AMP dataset}
\label{ampds_experiments}
\centering
\begin{tabular}{|c|c|c|c|}
\hline
\thead{Method} & \thead{Time-steps} 
    & \multicolumn{2}{c|}{\thead{Avg.MAPE \%}}
    \\ \cline{3-4} &&
	\multicolumn{1}{c|}{\thead{DeepDeFF}}
	&
	\multicolumn{1}{c|}{\thead{Basic}}\\


\hline
BGRU & 2 & 25.08 & 28.17 \\
\hline
GRU & 2  & 25.48 & 27.65 \\
\hline
BLSTM & 2 & 25.44 & 28.14 \\
\hline
LSTM & 2  & 25.57 & 28.57 \\
\hline
BRNN & 2 & 24.77 & 26.88 \\
\hline
\bfseries{RNN} & \bfseries{2}  & 25.62 & \bfseries{26.26} \\

\hline
\bfseries{BGRU} & \bfseries{6} & \bfseries{24.64} & 31.58 \\
\hline
GRU & 6 & 25.33 & 30.49 \\
\hline
BLSTM & 6 & 25.98 & 32.32 \\
\hline
LSTM & 6 & 26.00 & 33.47 \\
\hline
BRNN & 6 & 25.47 & 30.10 \\
\hline
RNN & 6 & 25.32 & 27.97 \\

\hline
BGRU & 12 & 25.19 & 32.85 \\
\hline
GRU & 12 & 25.17 & 30.00 \\
\hline
BLSTM & 12 & 26.62 & 36.08 \\
\hline
LSTM & 12 & 25.81 & 30.65 \\
\hline
BRNN & 12 & 25.35 & 31.62 \\
\hline
RNN & 12 & 25.87 & 29.54 \\
\hline
\end{tabular}
\end{table}

\begin{figure}[!htb]
    \centering
    \includegraphics[width=0.48\textwidth]{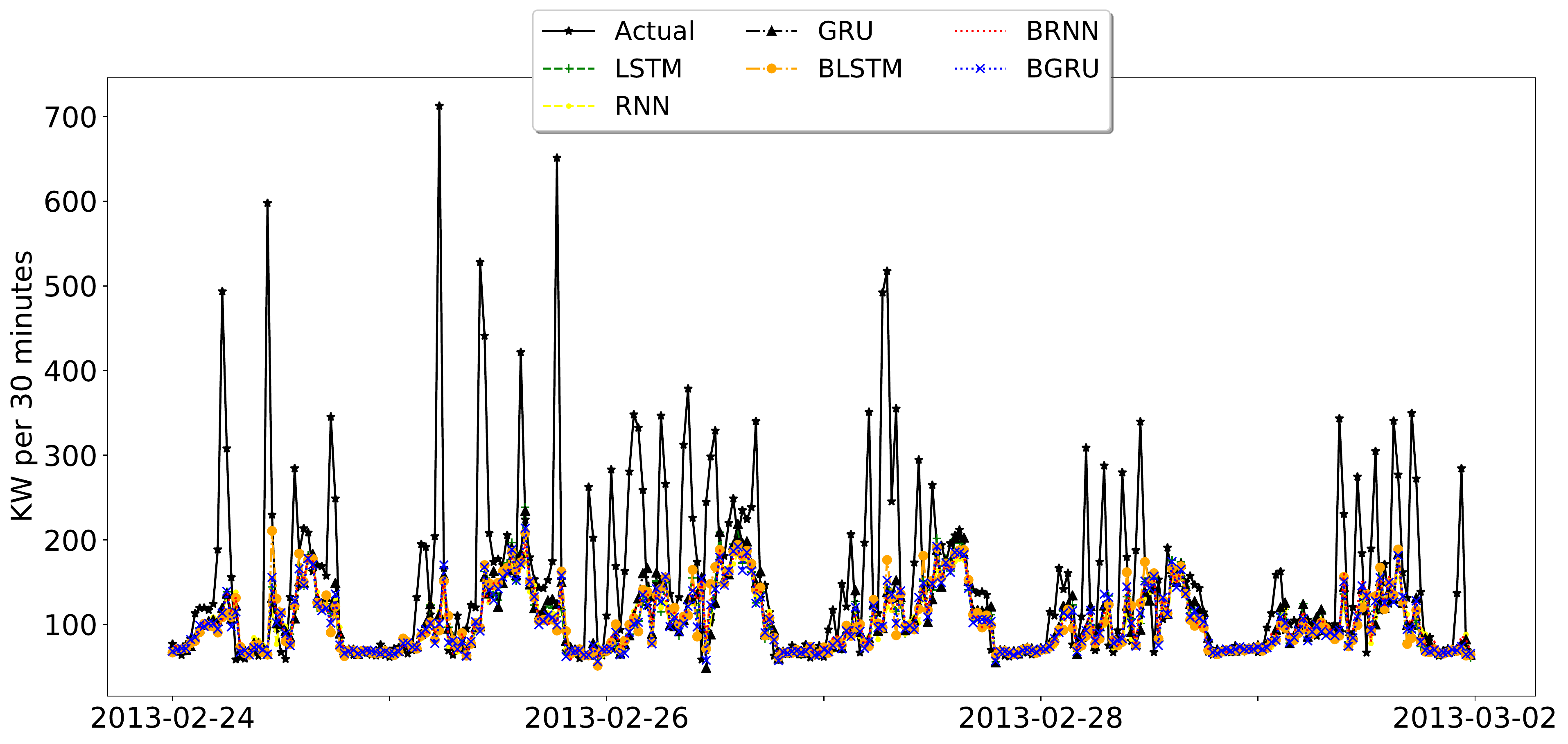}
    \caption{Actual load versus load predicted by DeepDeFF architecture utilizing different sequential layers.}
    \label{ampds_actual_vs_predicted}
\end{figure} 

Fig.~\ref{ampds_actual_vs_predicted} shows that the DeepDeFF architecture performs well in predicting the general load and suffers in case of outliers. This is because the model was able to learn the underlying general pattern from the training data, and gave it more importance than to outliers. This problem occurred because the training data was not enough and does not cover all the months; so the test data is of a month that was never seen during training.

\subsection{RTE Dataset}

\subsubsection{Train \& Test Setting}

RTE data is subdivided into three subsets with a split ratio of 0.7/0.2/0.1 as:

\begin{enumerate}[label=(\alph*)]
    \item Training set (01-Jan-2013 to 18-Nov-2015)
    \item Validation set (19-Nov-2015 to 07-Aug-2016)
    \item Test set (08-Aug-2016 to 31-Dec-2016)
\end{enumerate}

\begin{figure}[!htb] 
\centering
\includegraphics[width=0.48\textwidth]{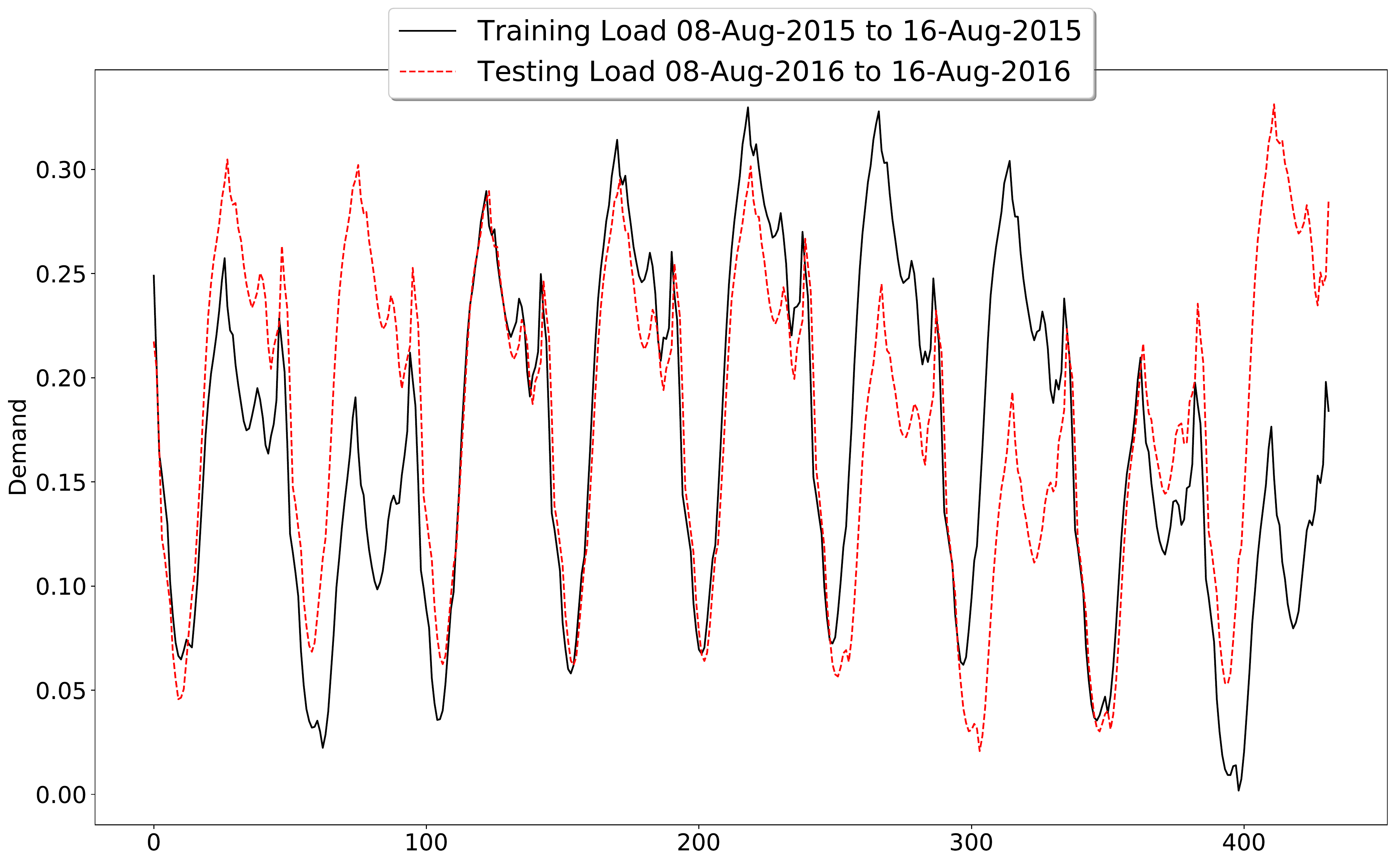}



\caption{Training and testing data comparison, RTE dataset}
\label{rte_train_vs_test}

\end{figure}

Fig.~\ref{rte_train_vs_test} show the subsets of training and testing data for the dates mentioned in the figures' legends. Such close resemblance in the test and train data helps the model to make accurate predictions as evident from the results.


\subsubsection{Results}
\begin{figure}[!htb]
    \centering
    \includegraphics[width=0.48\textwidth]{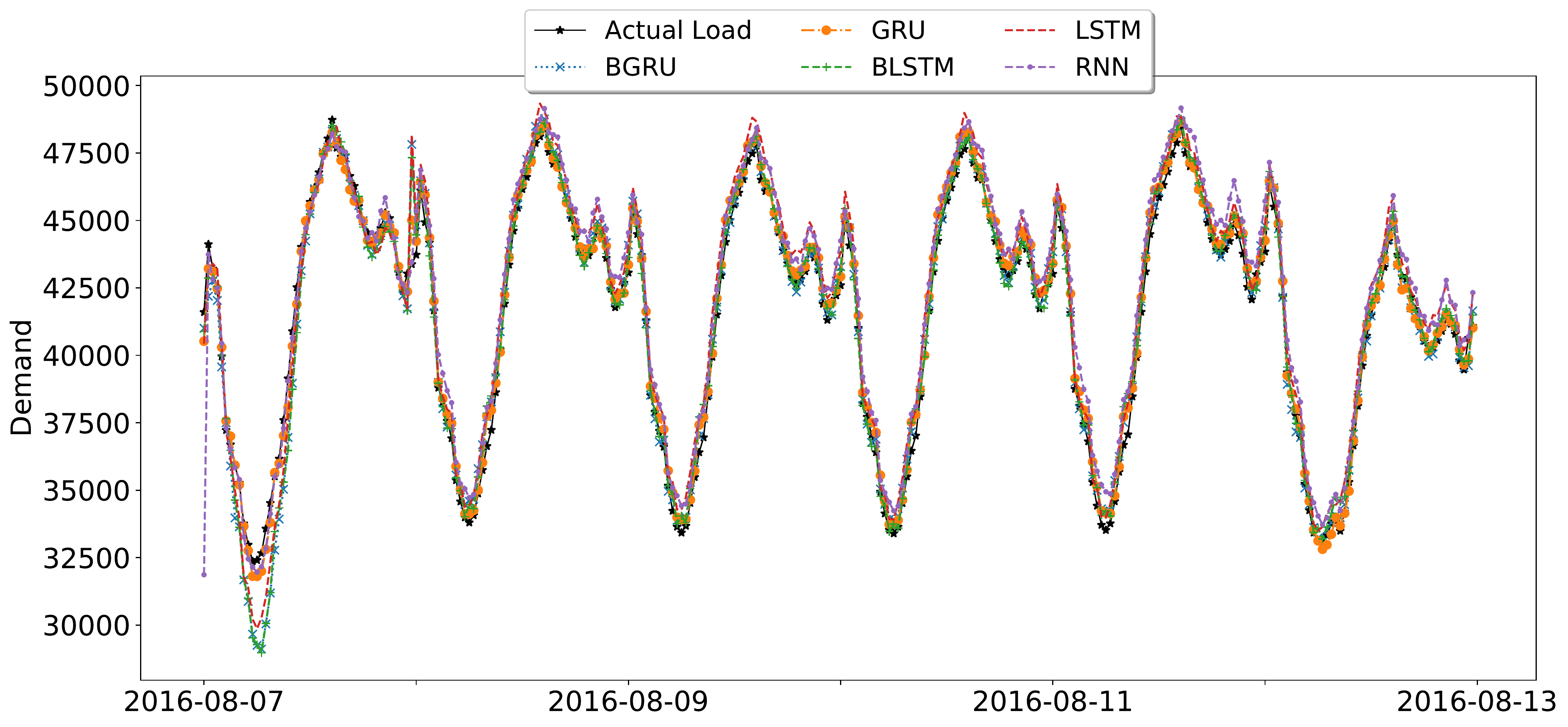}
    \caption{Actual load versus load predicted by DeepDeFF architecture utilizing different sequential layers.}
    \label{rte_actual_vs_predicted}
\end{figure}

It is observed from the results for SGSC and AMPds datasets that the experiments with 2 time-steps mostly yield the best results. Henceforth 2 time steps is used for the experiments on other datasets. Table-\ref{rte_experiments} shows the results for RTE dataset. The proposed model with GRU and derived features performed best with average MAPE of 0.81\%. Fig.~\ref{rte_actual_vs_predicted} shows the prediction results against the actual system load which further confirms the excellent performance of the DeepDeFF architecture.








\begin{table}[htb]
\renewcommand{\arraystretch}{1.2}

\caption{Results achieved on RTE dataset}
\label{rte_experiments}
\centering
\begin{tabular}{|c|c|c|c|}
\hline
\thead{Method} & \thead{Time-steps} 
    & \multicolumn{2}{c|}{\thead{Avg.MAPE \%}}
    \\ \cline{3-4} &&
	\multicolumn{1}{c|}{\thead{DeepDeFF}}
	&
	\multicolumn{1}{c|}{\thead{Basic}}
	\\

\hline
BGRU & 2 & 0.84& 1.06\\
\hline
\bfseries{GRU} & \bfseries{2} & \bfseries{0.81} & 1.09 \\
\hline
\bfseries{BLSTM} & \bfseries{2} & 1.63 & \bfseries{1.04} \\
\hline
LSTM & 2 & 1.17 & 1.19 \\
\hline
BRNN & 2 & 1.26 & 1.13 \\
\hline
RNN & 2 & 1.39 & 1.14 \\

\hline

\end{tabular}
\end{table}

\subsection{ERCOT Dataset}
\subsubsection{Train \& Test Setting}
ERCOT data is subdivided into three subsets with a split ratio of 0.5/0.1/0.4 as:

\begin{enumerate}[label=(\alph*)]
    \item Training set (01-Jan-2011 to 26-May-2013)
    \item Validation set (27-May-2013 to 31-Dec-2013)
    \item Test set (01-Jan-2014 to 31-Dec-2015)
\end{enumerate}






Similar to RTE, ERCOT is also the accumulated load consumption data of Texas. The train and test data for ERCOT also has close resemblance similar to Fig.~\ref{rte_train_vs_test}

\subsubsection{Results}
\begin{figure}[htb]
    \centering
    \includegraphics[scale=1,width=0.48\textwidth]{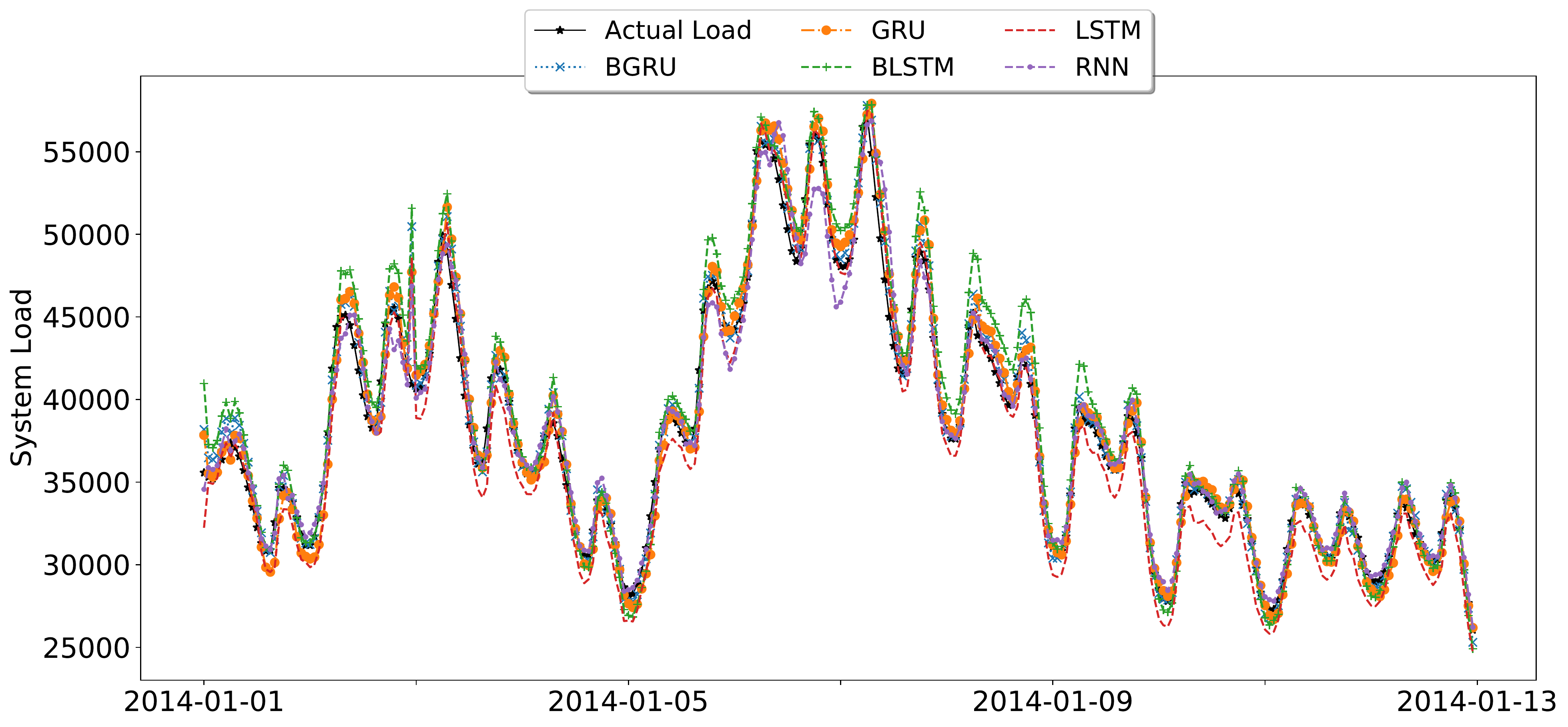}
    \caption{Actual load versus load predicted by DeepDeFF architecture utilizing different sequential layers.}
    \label{ercot_results}
\end{figure}

Table-IV shows the results for ERCOT dataset where the DeepDeFF model with BGRU performed best with average MAPE of 0.91\%. Figure-\ref{ercot_results} shows the results that establishes the effectiveness of the DeepDeFF architecture.

\begin{table}[htb]
\renewcommand{\arraystretch}{1.2}

\caption{Results achieved on ERCOT dataset}
\label{ercot_experiments}
\centering
\begin{tabular}{|c|c|c|c|}
\hline
\thead{Method} & \thead{Time-steps} 
    & \multicolumn{2}{c|}{\thead{Avg.MAPE \%}}
    \\ \cline{3-4} &&
	\multicolumn{1}{c|}{\thead{DeepDeFF}}
	&
	\multicolumn{1}{c|}{\thead{Basic}}
	\\
\hline
\bfseries{BGRU} & \bfseries{2} & \bfseries{0.91}& 1.40\\
\hline
GRU & 2 & 1.38 & 1.76 \\
\hline
BLSTM & 2 & 0.98 & 2.58 \\
\hline
LSTM & 2 & 1.17 & 1.96 \\
\hline
\bfseries{BRNN} & \bfseries{2} & 0.92 & \bfseries{1.37} \\
\hline
RNN & 2 & 1.38 & 2.88 \\

\hline
\end{tabular}
\end{table}








\subsection{PRECON}
\subsubsection{Train \& Test Setting}
Owing to the peculiar nature of the PRECON dataset, it is pre-processed in two steps in this research. First, the data is converted from 1-minute interval to 30-minute intervals by taking the average over the 30 consecutive load readings. The second step is to take care of close to zero values in the data that are mostly due to power outages. Otherwise these values cause divide-by-zero problem when using MAPE function for evaluation, resulting into unrealistically high MAPE and adversely effecting the performance of the machine learning algorithm. This is countered simply by adding a small offset of 0.1 KW to all the readings. The offset is small enough to makes no significant change in the nominal values of the load and takes effect only for the near zero data. This simplest pre-processing has shown remarkable impact on the performance of the DeepDeFF algorithm as evident from the results.

The data splitting is done in a unique way here due to the reason that it spanned over a period of only one year with no repeated data for any month. So instead of using an overall split of data, as done in previous datasets, a month-wise split is proposed. Here the training, validation, and testing data is taken from days 1 – 21, 22 – 26, 27 – 30/31 respectively for each month. This corresponds roughly to an overall split of 0.7/0.2/0.1.
%

\subsubsection{Results}
Table V shows the comparison of results obtained for PRECON dataset. DeepDeFF models have consistently outperformed the basic models on all the houses, achieving on average 8.9\% lower MAPE than basic models.

\begin{figure}[!htb] 

\begin{subfigure}{0.48\textwidth}
    \includegraphics[width=1\textwidth]{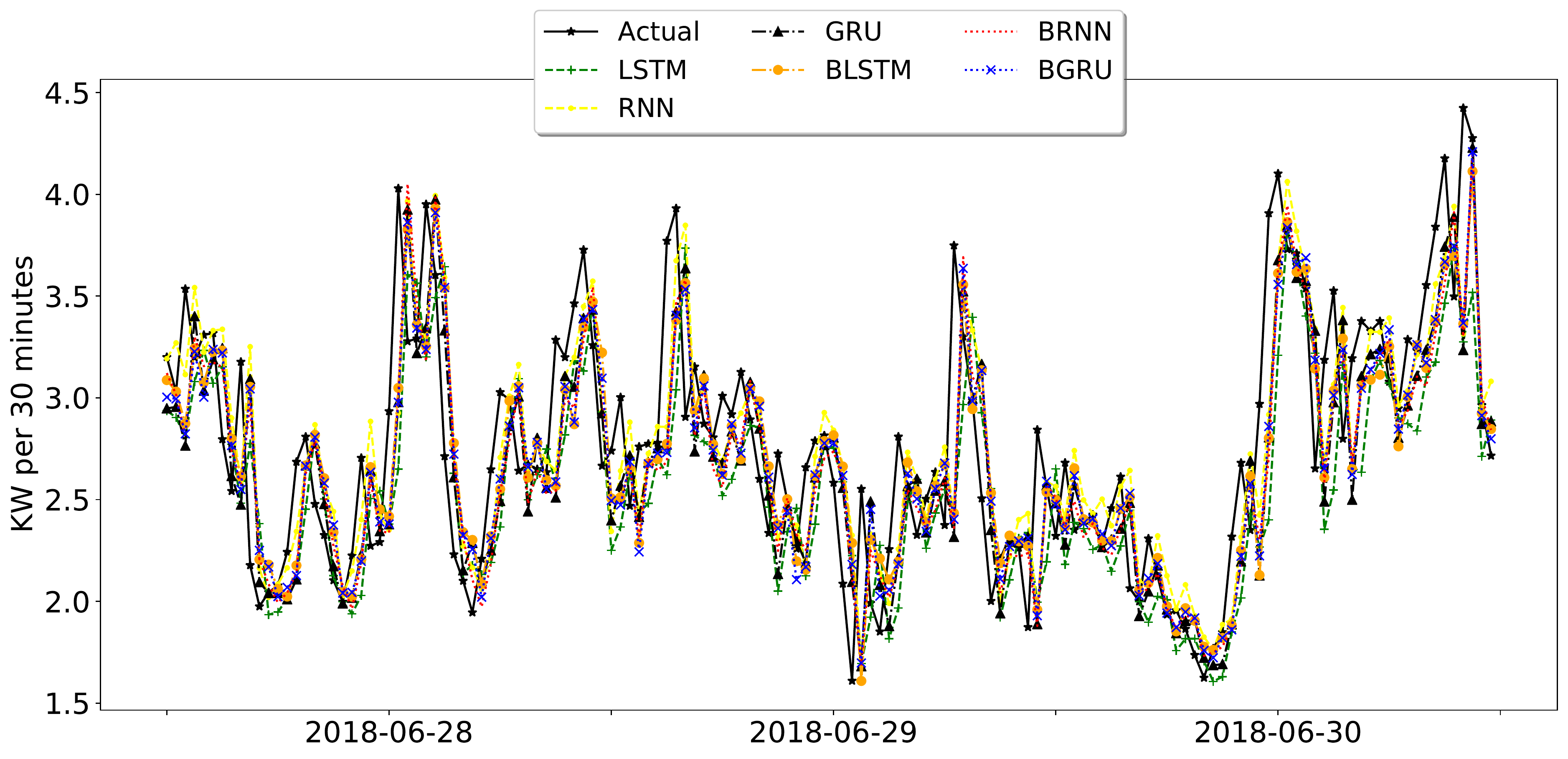}
    \caption{Prediction of DeepDeFF models vs actual load for House 3}
    \label{Figure Precon.A}

\end{subfigure}
%
\begin{subfigure}{0.48\textwidth}
    \includegraphics[width=1\textwidth]{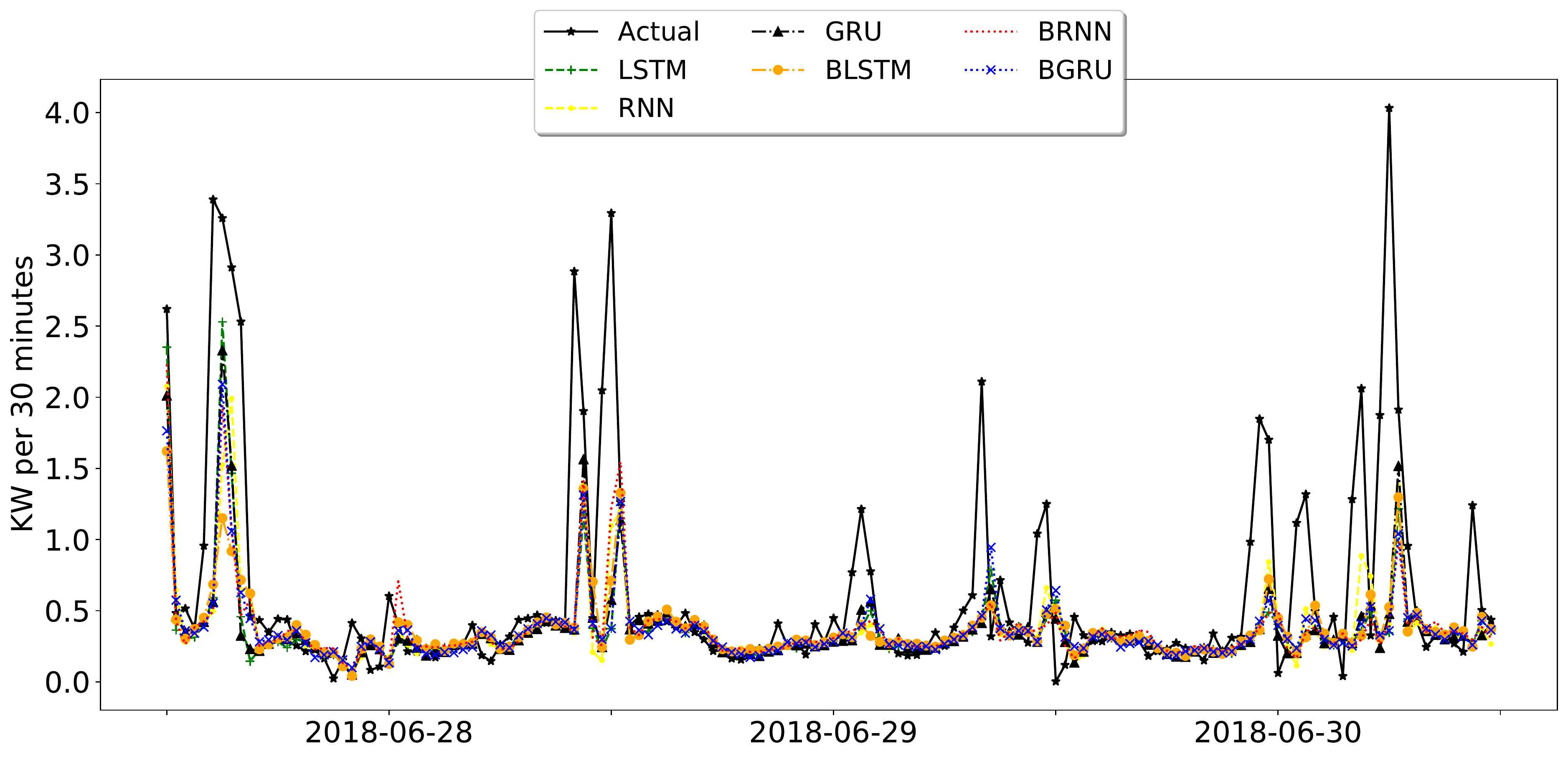}
    \caption{Prediction of DeepDeFF models vs actual load for House 29} 
    \label{Figure Precon.B}

\end{subfigure}
\caption{Actual load versus load predicted by DeepDeFF architecture utilizing different sequential layers.}
\label{Figure Precon}
\end{figure}

The value of the MAPE achieved by DeepDeFF models ranged from 7.67\% on House 3 to 37.61\% on House 29. The graphs of predicted versus actual load of these two houses are shown in Fig.~\ref{Figure Precon.A} and Fig.~\ref{Figure Precon.B} respectively.

\begin{table}[htbp]
\renewcommand{\arraystretch}{1.2}

\caption{Results achieved on PRECON dataset}
\label{precon_experiments}
\centering
\begin{tabular}{|c|c|c|c|c|c|}
\hline
\thead{Method} & \thead{Time-steps} 
    & \multicolumn{2}{c|}{\thead{Avg.MAPE \%}}
    \\ \cline{3-4} &&
	\multicolumn{1}{c|}{\thead{DeepDeFF}}
	&
	\multicolumn{1}{c|}{\thead{Basic}}
	\\

BGRU & 2  & 21.87 & 24.01 \\
\hline
GRU & 2  & 22.1 & 24.04 \\
\hline
BLSTM & 2 & 22 & 24.18\\
\hline
LSTM & 2  & 22.18 & 24.31 \\
\hline
\bfseries{BRNN} & \bfseries{2} & \bfseries{21.67} & \bfseries{23.9} \\
\hline
RNN & 2 & 21.89 & 24.11 \\

\hline

\end{tabular}
\end{table}

The above mentioned results on five public datasets infer that SLF for individual households: SGSC, AMPDs and PRECON is more difficult than aggregated load forecasting of a country or state wide dataset because of high variances in load consumption patterns of the former ones. However the proposed DeepDeFF architecture has been able to forecast better than the previously published techniques.


\section{Conclusion}
Load forecasting is of critical importance to optimally schedule and reliably manage the operations of power systems. This manuscript presented a deep learning architecture based on sequential layers, and a pre-processing method for introducing hand-crafted features into the end-to-end learning pipeline of the deep learning model, for short-term load forecasting. It is demonstrated with rigorous experimentation that the inclusion of hand-crafted features has improved the learning and predictions of the model, specially for smaller datasets. The proposed DeepDeFF architecture has been comprehensively tested on five different datasets -- two country/state wide datasets and three household datasets. The results achieved from the proposed methodology beat the current benchmark of these datasets for SLF.


\ifCLASSOPTIONcaptionsoff
  \newpage
\fi



%

\bibliographystyle{IEEEtran}
\bibliography{SLF_DeepDeFF.bib}

%








\end{document}